\documentclass[lettersize,journal]{IEEEtran}
\usepackage{amsmath,amsfonts}
\usepackage{algorithmic}
\usepackage{algorithm}
\usepackage{array}
\usepackage[caption=false,font=normalsize,labelfont=sf,textfont=sf]{subfig}
\usepackage{textcomp}
\usepackage{stfloats}
\usepackage{url}
\usepackage{verbatim}
\usepackage{graphicx}
\usepackage{cite}

\usepackage{algorithm}
\usepackage{algorithmic}
\usepackage{threeparttable}
\usepackage{bbding}
\usepackage{utfsym}
\usepackage{amsmath}
\usepackage{color}
\usepackage{xcolor}         
\usepackage{colortbl}
\definecolor{tabgray1}{gray}{.9}
\definecolor{tabgray2}{gray}{.84}
\usepackage{multirow} 
\usepackage{makecell}
\usepackage{pifont}
\definecolor{green}{rgb}{0.0,1.0,0.0}

\definecolor{redd}{rgb}{1.0,0.0,0.0}

\usepackage{bm}
\definecolor{mygray}{gray}{.7}
\definecolor{mypink}{rgb}{.99,.91,.95}
\definecolor{mycyan}{cmyk}{.3,0,0,0} 
\usepackage{calligra}
\usepackage{color}
\usepackage{amssymb}
\definecolor{BlueColor}{rgb}{0.0, 0.0, 1.0}
\definecolor{RedColor}{rgb}{1.0, 0.0, 0.0}
\definecolor{BlackColor}{rgb}{0.0, 0.0, 0.0}
\definecolor{GreenColor}{rgb}{0.3, 0.6, 0.4}

\usepackage{lineno}
\usepackage{booktabs}
\usepackage{xspace} 

\usepackage{newfloat}
\usepackage{listings}
\definecolor{evgreen}{HTML}{2ca02c}
\definecolor{evred}{HTML}{d62728}
\newcommand{\cmark}{\ding{51}} 
\newcommand{\xmark}{\ding{55}} 

\newcommand{\rev}[1]{{\color{black}#1}}

\usepackage{tabularx}
\newcolumntype{Y}{>{\centering\arraybackslash}X}
\newcolumntype{L}{>{\raggedright\arraybackslash}X}

\usepackage{bibentry}

\begin{document}
\title{Rethinking Event-Based Object Detection through Representation-Level Temporal Aggregation and Model-Level Hypergraph Reasoning}

\author{
Meisen Wang, Hao Deng, Wei Bao, Chengjie Wang, Zhiqiang Tian, Shaoyi Du, Siqi Li, and Yue Gao,~\IEEEmembership{Senior Member,~IEEE}
\thanks{Meisen Wang and Zhiqiang Tian are with the School of Software Engineering, Xi'an Jiaotong University, Xi'an 710049, China (e-mail: 3125158005@stu.xjtu.edu.cn; zhiqiangtian@xjtu.edu.cn).}
\thanks{Hao Deng, Zhiqiang Tian, and Shaoyi Du are with the National Key Laboratory of Human-Machine Hybrid Augmented Intelligence, the National Engineering Research Center for Visual Information and Applications, and the Institute of Artificial Intelligence and Robotics, Xi'an Jiaotong University, Xi'an 710049, China (e-mail: denghao293@stu.xjtu.edu.cn; zhiqiangtian@xjtu.edu.cn; dushaoyi@xjtu.edu.cn).}
\thanks{Wei Bao, Siqi Li, and Yue Gao are with \{BNRist, THUIBCS, BLBCI, School of Software\}, Tsinghua University, Beijing 100084, China, and also with the Yangtze Delta Region Institute, Tsinghua University, Jiaxing 314006, China (e-mail: \{baoweivvv, lisiqi19971013, kevin.gaoy\}@gmail.com).}
\thanks{Chengjie Wang is with the College of Grassland Science, Inner Mongolia Agricultural University, Hohhot 010018, China (e-mail: nmgcjwang3@imau.edu.cn).}
}

\maketitle

\begin{abstract}

Event cameras provide microsecond-level temporal resolution, low latency, and high dynamic range, offering potential for perception under fast motion and challenging illumination conditions. However, existing Event-based Object Detection (EOD) methods face limitations at both the representation and model levels: prior event representations usually encode temporal information indirectly through redundant structures, while detection models struggle to explicitly aggregate fragmented event responses into coherent high-order object features. To address these limitations, we present \textbf{Event Dual Temporal-Relational Aggregation Detector (Ev-DTAD)}, a unified EOD framework that integrates representation-level temporal encoding with model-level temporal-hypergraph reasoning. Specifically, we introduce \textbf{Hierarchical Temporal Aggregation (HTA)}, a compact three-channel pseudo-RGB representation that explicitly embeds temporal information across intra- and inter-window events. To further enhance detection under sparse and fragmented event responses, we propose \textbf{Frequency-aware Hypergraph Temporal Fusion (FHTF)}, which refines multi-scale event features through temporal evolution modeling and high-order relational reasoning. Extensive experiments on Gen1 (\textbf{+0.8 mAP}), 1Mpx/Gen4 (\textbf{+0.5 mAP}), and eTraM (\textbf{+3.0 mAP}) demonstrate that Ev-DTAD achieves a competitive accuracy--efficiency trade-off, validating the complementarity between compact temporal representation and temporal-hypergraph feature reasoning (Fig.~\ref{fig:bubble}).
The code is available at: \url{https://github.com/meisenwang/Ev-DTAD}.
\end{abstract}

\section{Introduction}
\label{sec:intro}
Event cameras output asynchronous streams of brightness changes with microsecond-level temporal resolution, low latency, and high dynamic range, making them particularly attractive for perception in fast motion and challenging illumination conditions~\cite{jiang2023event,lichtsteiner2008temporal,brandli2014davis,gallego2022event}. These advantages have motivated increasing interest in event-based perception for autonomous driving and robotics, spanning object detection~\cite{bao2026rm,maqueda2018event,detournemire2020large}, object tracking~\cite{zhu2025crsot}, and semantic segmentation~\cite{xie2024eisnet}. However, unlike conventional images~\cite{guo2026phaselora}, event streams are sparse, polarity-dependent, temporally fragmented, and susceptible to background activity noise, particularly under low-contrast conditions~\cite{gallego2022event,rebecq2021high,ding2023mlb}. Object-related event responses are often distributed across discontinuous events rather than captured in a complete spatial snapshot, making it challenging to preserve temporal structures and object features in a compact and informative representation~\cite{sironi2018hats,gehrig2019end,rebecq2021high}.

\begin{figure}[t]
\begin{center}
\includegraphics[width=\linewidth]{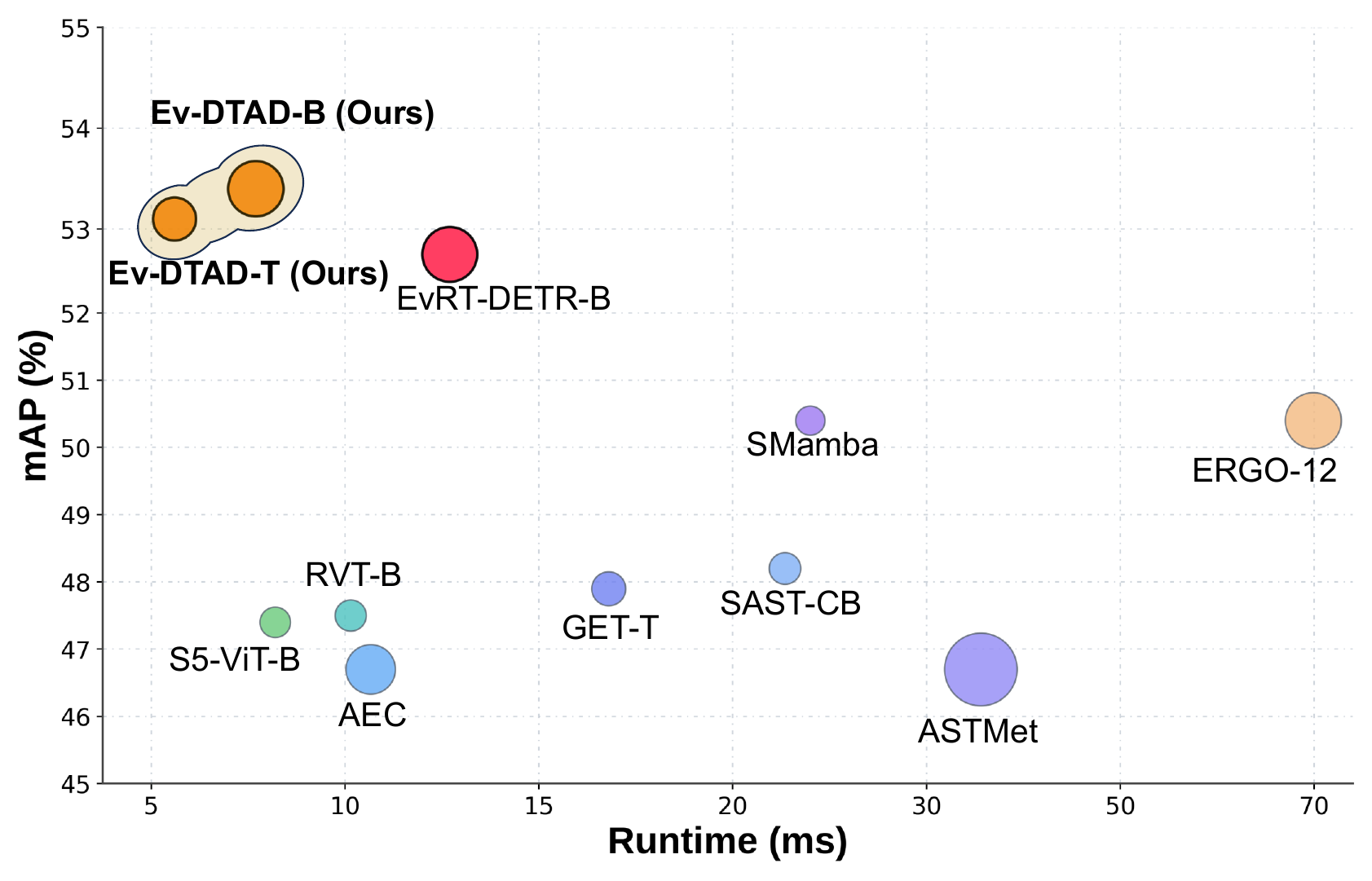}
\end{center}
\vspace{-0.3cm}
  \caption{\textbf{Latency--accuracy comparison on Gen1.} Bubble area is proportional to the number of parameters. Our models achieve state-of-the-art accuracy while maintaining competitive inference speed.}
  \label{fig:bubble}
\end{figure}

Existing EOD studies can be grouped into two relatively independent research lines: \textit{event representation} and \textit{detection model design}. \textit{(i) Event representation} aims to transform asynchronous events into more informative structures for effective detection~\cite{sironi2018hats,cannici2020matrix,li2022retinomorphic,peng2023better,wang2023dmanet,zubic2023chaos}. While these methods enhance access to event information, they capture temporal information indirectly through expanded temporal or feature dimensions~\cite{zhu2019unsupervised,gehrig2019end,peng2023get}, without explicitly embedding local event ordering or motion continuity into the representation, leading to redundancy. \textit{(ii) Detection model design} focuses on strengthening the detector itself by modeling temporal dependencies over event-derived features~\cite{li2022asynchronous,gehrig2023recurrent,peng2024scene,zubic2024state}. More broadly, graph-based event models have demonstrated the benefit of uncovering local and global spatiotemporal correlations among sparse events~\cite{alkendi2024neuromorphic}. These methods improve temporal feature propagation and help accumulate sparse evidence over time, but they are less explicit in structuring discontinuous event responses into high-order object-level associations. This limitation is particularly pronounced in EOD, where object evidence is typically sparse and fragmented, arising from motion-induced brightness changes rather than dense region-level observations~\cite{gallego2022event}.

Motivated by these observations, we present \textbf{Event Dual Temporal-Relational Aggregation Detector (Ev-DTAD)}, a unified EOD framework that bridges event representation and detection model design through two complementary levels. Ev-DTAD compactly preserves discriminative event dynamics at the representation level and strengthens high-order feature association over sparse event responses at the model level. It introduces two key components: \textbf{Hierarchical Temporal Aggregation (HTA)} for compact temporal event encoding, and \textbf{Frequency-aware Hypergraph Temporal Fusion (FHTF)} for temporal-relational feature refinement.

\begin{figure*}
\begin{center}
\includegraphics[width=0.95\linewidth]{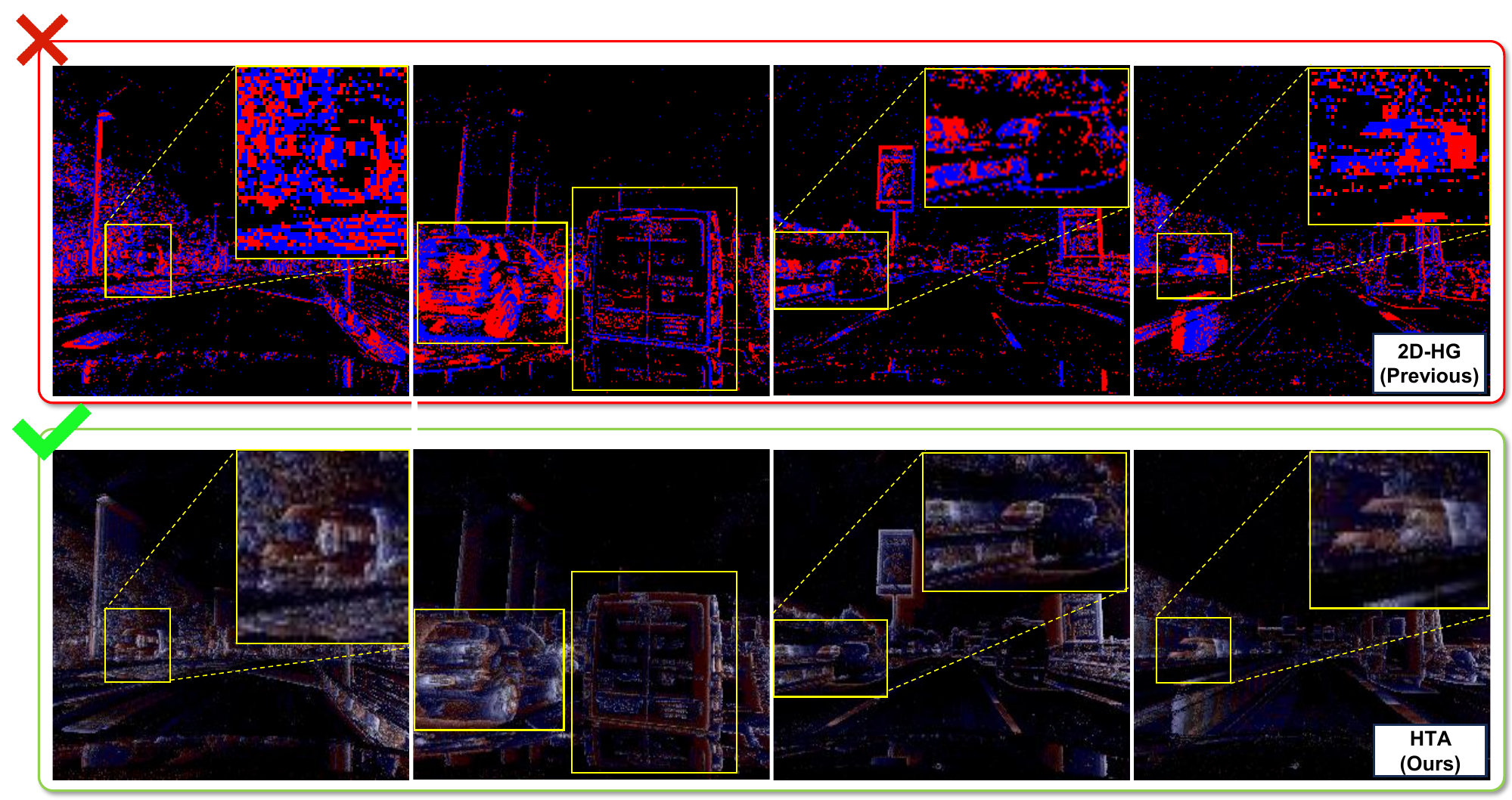}
\end{center}
\vspace{-0.5cm}
\caption{
    \textbf{Illustrative comparison between 2D Histogram (2D-HG) and Hierarchical Temporal Aggregation (HTA) representation.} 
    HTA is one of our key contributions, encoding temporal event information into a compact pseudo-RGB format. Compared with 2D-HG, HTA produces cleaner object structures with reduced background noise. Yellow boxes highlight representative differences.}
\vspace{-0.5cm}
\label{fig:comparsion2}
\end{figure*}

The first component, \textbf{HTA}, addresses the compactness--expressiveness trade-off by encoding asynchronous events into a compact pseudo-RGB representation while preserving discriminative structures. Unlike prior methods that merely accumulate events or expand along temporal dimensions, HTA explicitly embeds temporal information into representation. Event information are aggregated at two temporal scales: intra-window aggregation captures local ordering and recent motion cues, while inter-window aggregation propagates reliable historical responses across adjacent windows, maintaining temporal continuity. The resulting three-channel representation encodes both instantaneous and historical temporal dynamics, preserving foreground motion structures and suppressing background noise, as illustrated in Fig.~\ref{fig:comparsion2}.

The second component, \textbf{FHTF}, addresses the fragmented nature of event features at the model level. Although HTA provides compact temporal encoding, object evidence in EOD remains sparse, incomplete, and distributed across time and scales. FHTF refines multi-scale event features by combining stateful temporal modeling with frequency-aware hypergraph reasoning. The temporal branch propagates feature states across consecutive representations, while the hypergraph branch organizes discontinuous event responses into high-order structural contexts. Together, HTA and FHTF enable Ev-DTAD to combine compact event encoding with robust temporal-relational reasoning for event-based object detection.

We conduct experiments on Gen1~\cite{detournemire2020large}, 
1Mpx/Gen4~\cite{perot2020learning}, 
and eTraM~\cite{verma2024etram}. 
As illustrated in Fig.~\ref{fig:bubble}, Ev-DTAD establishes a competitive accuracy--efficiency trade-off, outperforming the strongest compared baselines in accuracy while reducing inference latency. 
These results validate the effectiveness of two-level EOD modeling for accurate and efficient EOD.
\begin{itemize}
    \item We propose \textbf{HTA}, a compact three-channel event representation that hierarchically organizes intra-window motion ordering and inter-window reliable responses into a detector-friendly pseudo-RGB format, reducing redundant temporal encoding and explicitly embedding temporal dynamics at the representation level while preserving foreground structures.
    
    \item We propose the \textbf{FHTF} module, a frequency-aware hypergraph temporal fusion module that performs high-order temporal-relational reasoning across event features at the model level, enhancing temporal robustness for sparse and temporally fragmented event streams.

    \item We build \textbf{Ev-DTAD}, a two-level EOD framework that achieves a strong accuracy--efficiency trade-off on Gen1 (+0.8 mAP), 1Mpx/Gen4 (+0.5 mAP), and eTraM (+3.0 mAP), demonstrating the complementarity between compact temporal representation and feature-level temporal-relational refinement.
\end{itemize}

\section{Related Work}

\subsection{Event Representation.}
A common strategy in EOD is to convert asynchronous event streams into dense grid-based representations, enabling image-style detection networks to process event data efficiently~\cite{gallego2022event,perot2020learning,gehrig2023recurrent,yang2025smamba,peng2024scene}. Conventional representations, such as event histograms, time surfaces, and voxel grids, accumulate event counts, timestamps, or polarities within fixed temporal windows~\cite{sironi2018hats,zhu2019unsupervised,gehrig2019end}. Recent methods preserve richer spatiotemporal information through learned or multi-channel conversion. Matrix-LSTM~\cite{cannici2020matrix} learns spatial event memories, Peng et al.~\cite{peng2023better} introduce Hyper Histogram and Adaptive Event Conversion to encode polarity and temporal features, and Zubić et al.~\cite{zubic2023chaos} propose ERGO-12 by optimizing the selection and ordering of representation components. Although these methods demonstrate the importance of representation design, richer temporal information is generally retained through additional bins, features, or channels, resulting in a trade-off between compactness and temporal expressiveness. In contrast, HTA explicitly organizes local event ordering and cross-window motion continuity within a compact three-channel representation.

\subsection{Temporal Modeling in Event-Based Object Detection.}
Temporal modeling is another major direction in EOD, aiming to accumulate sparse object evidence across consecutive event windows. Early recurrent detectors such as RED~\cite{perot2020learning} introduce temporal memory, while ASTM-Net~\cite{li2022asynchronous} adopts explicit memory for continuous aggregation. DMANet~\cite{wang2023dmanet} combines long- and short-term memory, RVT~\cite{gehrig2023recurrent} employs recurrent transformers, and GET~\cite{peng2023get} organizes events into structured tokens to model spatial, temporal, and polarity interactions. Later methods improve temporal modeling through temporal consistency learning~\cite{wu2024leod}, scene-adaptive sparse attention~\cite{peng2024scene}, state-space modeling~\cite{zubic2024state}, or detector adaptation~\cite{torbunov2025evrtdetr}. Beyond dense-grid modeling, AEGNN~\cite{schaefer2022aegnn} processes events as evolving spatio-temporal graphs, demonstrating the potential of structural reasoning over sparse events. However, existing methods primarily emphasize temporal propagation or pairwise feature interactions, while the joint modeling of temporal evolution, frequency characteristics, and high-order event-feature associations remains insufficiently explored. Our FHTF addresses this limitation by integrating stateful temporal fusion with frequency-aware hypergraph reasoning.

\subsection{Hypergraph-Based High-Order Reasoning.}
Hypergraphs can connect multiple nodes through a single hyperedge and are therefore well suited to modeling high-order relations beyond pairwise interactions. Hypergraph neural networks have been extensively studied through spectral convolution, dynamic construction, and node--hyperedge--node message passing~\cite{feng2019hypergraph,yadati2019hypergcn,jiang2019dynamic,bai2021hypergraph,huang2021unignn,chien2021you}. In computer vision, hypergraph-based methods capture contextual and structural dependencies among visual regions~\cite{li2013contextual,han2023vision}, while recent object detection studies demonstrate their effectiveness in modeling high-order correlations among visual features~\cite{feng2024hyper}. M$^{2}$C-EvDet~\cite{bao2026rm} further transfers frequency-decoupled semantics and hypergraph-based high-order relations from RGB images to event detectors through cross-modal distillation. However, its hypergraph reasoning primarily serves knowledge transfer during training, whereas high-order temporal refinement within event-only features remains underexplored. Combined with the effectiveness of frequency-domain modeling in visual representation learning~\cite{xu2020learning,chi2020fast,qin2021fcanet,rao2021global,chen2024fadc}, this motivates FHTF to construct frequency-aware high-order associations directly among temporally evolving event features.

\begin{figure*}
\begin{center}
\includegraphics[width=1.0\linewidth]{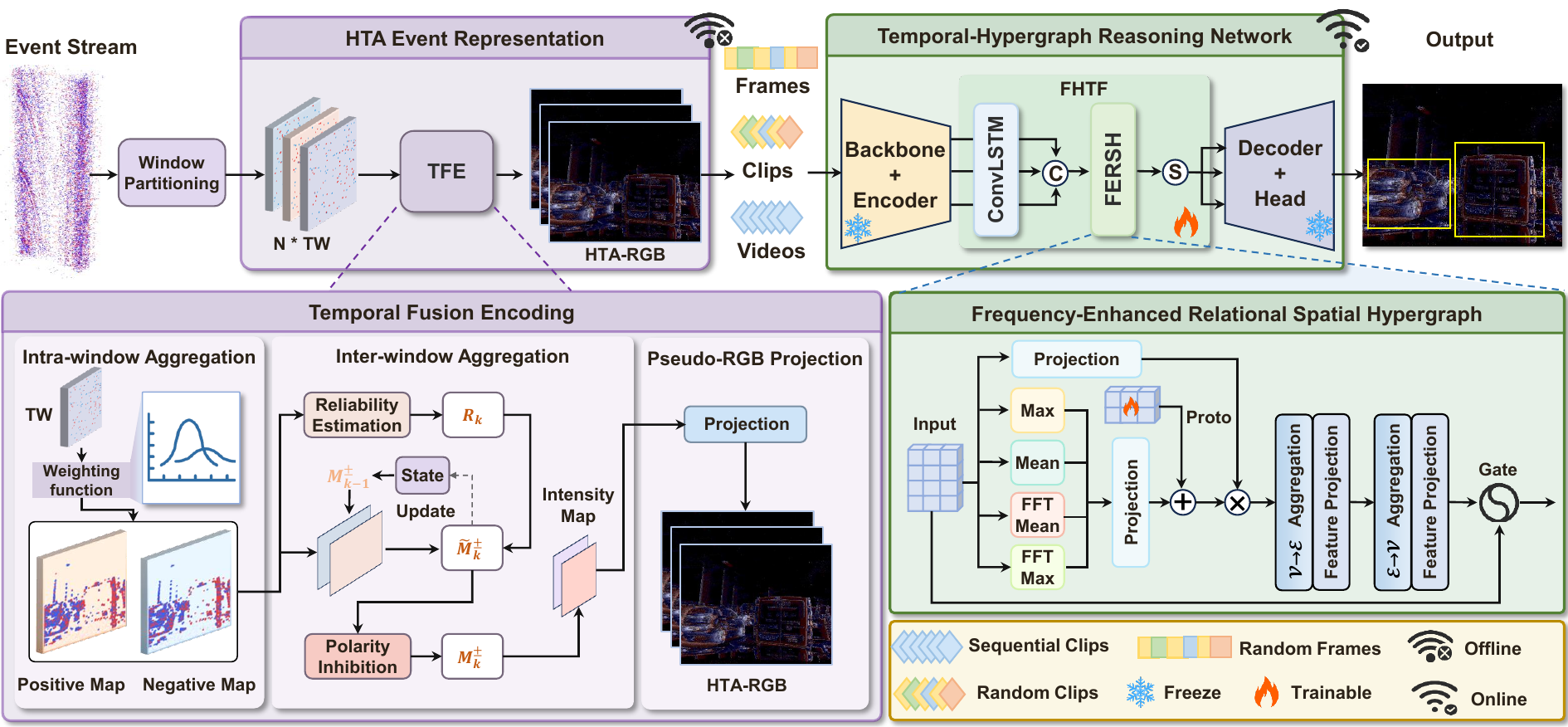}
\end{center}
\vspace{-0.5cm}
    \caption{\textbf{Framework overview.} Ev-DTAD consists of two core components: offline HTA representation generation and the FHTF module. It first converts asynchronous events into compact HTA frames offline. Consecutive frames are grouped into clips/videos and fed into the network, where multi-scale features are extracted and refined by FHTF through temporal evolution and frequency-aware hypergraph reasoning before query-based object prediction. \rev{TW denotes an event window, FFT the fast Fourier transform, and Proto learnable hyperedge prototypes.}}
\vspace{-0.5cm}
\label{fig:pipeline}
\end{figure*}

\section{Method}

This section presents Ev-DTAD, a unified framework for EOD that combines representation-level temporal encoding with model-level temporal-hypergraph reasoning. We first introduce HTA, a compact event representation generated through hierarchical temporal aggregation, in Sec.~\ref{sec:hta}. We then describe FHTF, which refines multi-scale event features through recurrent temporal propagation and frequency-aware hypergraph reasoning, in Sec.~\ref{sec:FHTF}. Finally, Sec.~\ref{sec:evdtad} summarizes how the above two components are integrated into the complete Ev-DTAD detection pipeline, as illustrated in Fig.~\ref{fig:pipeline}.

\subsection{HTA Event Representation}
\label{sec:hta}
To explicitly encode temporal structures at the representation level, the asynchronous event stream is converted into a compact pseudo-RGB representation through hierarchical temporal aggregation across intra- and inter-window event dynamics. The former preserves relative event timing within each window, whereas the latter maintains reliable motion responses across adjacent windows. Formally, an event stream is represented as
\begin{equation}
\mathcal{E}=\{e_i\}_{i=1}^{N}, \qquad e_i=(x_i,y_i,t_i,p_i),
\end{equation}
where $x_i$ and $y_i$ denote the spatial coordinates, $t_i$ denotes the timestamp, and $p_i\in\{-1,+1\}$ denotes the event polarity. We divide the asynchronous event stream into consecutive temporal windows $\mathcal{W}_k=\{e_i \mid T_k \le t_i < T_k+\Delta T\}$ of fixed length $\Delta T$.

\textbf{Intra-window temporal aggregation.} For each event $e_i\in\mathcal{W}_k$, a temporal weight is assigned as
\begin{equation}
\omega_i=\lambda+(1-\lambda)\left(\frac{t_i-T_k}{\Delta T}\right)^{\gamma_t},
\end{equation}
where $\lambda\in[0,1]$ prevents early events from being fully suppressed, and $\gamma_t$ controls the temporal recency bias. Events closer to the window endpoint consequently receive larger weights and better reflect the most recent object state. Positive and negative events are then separately accumulated with these weights to obtain window-level response maps $P_k$ and $N_k$, preserving event activity and coarse temporal ordering compactly.

\textbf{Reliability-aware inter-window temporal aggregation.} To propagate temporal information across adjacent windows, two polarity-specific state maps, $M_k^+$ and $M_k^-$, are maintained. Before updating them, we estimate the reliability of the current window response based on local activity intensity and polarity consistency. Using a local averaging operator $\mathcal{B}(\cdot)$, the reliability map is defined as
\begin{equation}
R_k=
\mathrm{clip}\!\left(
\frac{\mathcal{B}(A_k)}{\mathcal{B}(A_k)+\tau}
\cdot
\frac{|\mathcal{B}(S_k)|}{\mathcal{B}(A_k)+\varepsilon},
\,0,\,1
\right),
\end{equation}
where $A_k=P_k+N_k$ and $S_k=P_k-N_k$. Here, $\tau$ is the activity normalization constant and $\varepsilon$ is a small value to avoid division by zero. The two factors jointly favor locally active regions with a consistent polarity tendency while reducing the influence of weak or conflicting responses.

Based on the local reliability, an adaptive decay rate is defined as
\begin{equation}
\kappa_k=\mathrm{clip}\!\left(\kappa_0\bigl(1+\alpha(1-R_k)\bigr),\,\kappa_{\min},\,\kappa_{\max}\right),
\end{equation}
and update the two state maps by
\begin{equation}
\widetilde{M}_k^{\pm}=(1-\kappa_k\Delta T)^{b}\odot M_{k-1}^{\pm}+c\,E_k^{\pm},
\end{equation}
where $E_k^{+}=P_k$ and $E_k^{-}=N_k$. $\kappa_0$ is the base decay rate, $\alpha$ is the modulation coefficient, $b$ is the decay exponent, $c$ is the event injection coefficient, and $\kappa_{\min}$ and $\kappa_{\max}$ are the minimum and maximum decay rates used to bound the decay strength. This adaptive update retains reliable historical structures while accelerating the forgetting of uncertain responses.

To suppress simultaneous responses of opposite polarities at the same
location, polarity-competitive inhibition is applied, followed by smooth
saturation to bound the state magnitude:
\begin{equation}
\begin{aligned}
M_k^{+}
&= C\tanh\!\left(
\frac{\max\!\left(
\widetilde{M}_k^{+}-\beta\widetilde{M}_k^{-},\,0
\right)}{C}
\right), \\
M_k^{-}
&= C\tanh\!\left(
\frac{\max\!\left(
\widetilde{M}_k^{-}-\beta\widetilde{M}_k^{+},\,0
\right)}{C}
\right),
\end{aligned}
\end{equation}
where $\beta$ is the inhibition coefficient and $C$ is the state cap.

\textbf{Pseudo-RGB projection of temporally aggregated states.} After the above two-stage temporal aggregation, the final polarity states are converted into a pseudo-RGB image. A structural intensity map and a polarity bias map are first defined as:
\begin{equation}
\begin{aligned}
\Pi_k
&= \frac{M_k^+-M_k^-}
        {M_k^++M_k^-+\varepsilon}, \\
U_k
&= (1-\eta)\max(M_k^+,M_k^-)
   + \eta\left|M_k^+-M_k^-\right|.
\end{aligned}
\end{equation}
A logarithmic compression is then applied to obtain the luminance term
\begin{equation}
Y_k
= \frac{\log(1+g\,U_k)}
       {\log(1+g\,\sigma)}.
\end{equation}
The final three-channel output is computed as
\begin{equation}
I_k =
\left[
\begin{aligned}
&\mathrm{clip}\!\left(
Y_k+\mu\max(\Pi_k,0),\,0,\,1
\right), \\
&Y_k, \\
&\mathrm{clip}\!\left(
Y_k+\mu\max(-\Pi_k,0),\,0,\,1
\right)
\end{aligned}
\right]^{\gamma_c}.
\end{equation}
where $g$ is the luminance gain, $\sigma$ is the normalization scale, $\eta$ balances dominant response and polarity difference, $\mu$ controls color strength, and $\gamma_c$ denotes the gamma parameter. \rev{The luminance term $Y_k$ provides a common structural component for all three channels. Before gamma correction, the green channel contains only $Y_k$, while the red and blue channels additionally encode positive and negative polarity dominance, respectively. Accordingly, reddish and bluish tints indicate the dominant polarity of the accumulated responses, whereas balanced polarities yield an achromatic appearance. Brightness reflects temporally aggregated structural intensity; the colors encode event polarity rather than natural scene color.}
After 8-bit quantization, $I_k\in\mathbb{R}^{H\times W\times 3}$ denotes the pseudo-RGB representation for the $k$-th window.

Overall, HTA forms a temporally integrated state through hierarchical temporal aggregation rather than an isolated single-window snapshot, jointly preserving recent event responses, polarity-aware motion cues, and cross-window continuity in a compact three-channel representation.

\subsection{FHTF: Frequency-aware Hypergraph Temporal Fusion}
\label{sec:FHTF}

To fully exploit the multi-scale spatiotemporal dynamics and high-order dependencies inherent in asynchronous event streams, the \textbf{FHTF} module is proposed. It refines the multi-scale feature pyramid $F_t = \{f_{t,i}\}_{i=1}^3$ by combining ConvLSTM-based local stateful temporal evolution with Frequency-Enhanced Relational Spatial Hypergraph. The former propagates object evidence over time, while the latter associates spatially distributed responses through high-order relations, producing temporally coherent, relation-enhanced, and structurally robust event features before prediction.

\textbf{Stateful Temporal Evolution.}  At each scale $i$, FHTF maintains a recurrent state to model the temporal evolution of event features without heavy spatiotemporal operators. To preserve pretrained spatial priors while injecting temporal dynamics, a residual gating mechanism is used:
\begin{equation}
    \hat{f}_{t,i} = f_{t,i} + \alpha_i \cdot \Psi_i(f_{t,i}, \mathcal{M}_{t-1,i}),
\end{equation}
where $\Psi_i$ denotes the recurrent transition function, $\mathcal{M}_{t-1,i}$ represents the hidden memory state propagated from the previous frame, and $\alpha_i$ is a learnable gating parameter initialized to zero. The residual path preserves the spatial representation learned during single-frame training, whereas the gate progressively controls the contribution of recurrent information. This formulation ensures that temporal context is integrated seamlessly only where it provides discriminative temporal information.

\textbf{Frequency-Enhanced Relational Spatial Hypergraph (FERSH).} A key component of FHTF is the frequency-aware hypergraph refinement stage. Unlike pairwise self-attention, FHTF constructs latent hyperedges to capture high-order semantic and structural relations.

The hypergraph is defined over the spatially aligned multi-scale nodes $X \in \mathbb{R}^{N \times C}$. \rev{Event responses induced by brightness changes often form sparse and fragmented structures. We use spectral amplitude statistics to supplement spatial pooling with complementary global information about feature variation, enabling hyperedge anchors to adapt to the characteristics of the current input.} Specifically, a Fast Fourier Transform (FFT) is applied over the feature sequence to extract spectral amplitude statistics:
\begin{equation}
    c_{\text{spec}} = [\text{mean}(|\mathcal{F}(X)|), \text{max}(|\mathcal{F}(X)|)].
\end{equation}
\rev{The mean and maximum amplitudes summarize the average and strongest spectral responses, respectively. Together with spatial global pooling statistics, these descriptors modulate the learnable prototype matrix $P \in \mathbb{R}^{H \times d_k}$ to generate $H$ input-conditioned hyperedge anchors. The affinities between spatial node features and these anchors determine the soft incidence matrix $\mathbf{H} \in \mathbb{R}^{N \times H}$, which specifies the contribution of each node to each hyperedge. Nodes with strong associations to the same conditioned anchor are jointly aggregated into a shared hyperedge representation. Spectral descriptors therefore influence high-order associations through anchor modulation, while the subsequent message passing operates on spatial node features.}FHTF performs dual-hop message passing by aggregating node features into hyperedges for global context and broadcasting enriched hyperedge representations back to nodes for high-order refinement:
\begin{equation}
    X_{\text{FHTF}} = X + \gamma \odot \Phi_{\text{broadcast}} \left( \mathcal{H} \cdot \Phi_{\text{agg}}(\mathcal{H}^{\top} X) \right),
\end{equation}
where $\Phi_{\text{agg}}$ and $\Phi_{\text{broadcast}}$ denote the aggregation and redistribution projections, respectively, and $\gamma$ serves as a channel-wise refinement gate. Through this frequency-aware hypergraph reasoning, the FHTF module emphasizes structurally informative frequency patterns and improves the coherence of object-related event features.

\begin{table*}[t!]
    \centering
    \caption{\textbf{Comparison on Gen1 and 1Mpx/Gen4.}
    Colors denote event representations. All runtimes are measured on a single NVIDIA RTX 3090 GPU in ms/frame.
    Best and second-best results are marked in \textbf{bold} and \underline{underline}, respectively. \textcolor{red}{Runtime excludes representation generation.}}
    \label{tab:main_results}
    \scriptsize
    \setlength{\tabcolsep}{3pt}
    \resizebox{\textwidth}{!}{
    \begin{tabular}{cccccccc}
        \toprule
        \multirow{2}{*}{\textbf{Method}} & \multirow{2}{*}{\textbf{Venue}} & \multirow{2}{*}{\textbf{Representation}} & \multicolumn{2}{c}{\textbf{Gen1}} & \multicolumn{2}{c}{\textbf{1Mpx/Gen4}} & \multirow{2}{*}{\textbf{Params (M)}} \\
        \cmidrule(lr){4-5} \cmidrule(lr){6-7}
         &  &  & \textbf{mAP (\%)} & \textbf{Runtime (ms)} & \textbf{mAP (\%)} & \textbf{Runtime (ms)} &  \\
        \midrule
        \rowcolor{blue!3}
        ASTMNet \cite{li2022asynchronous} & TIP'22 & Asynchronous Events & 46.7 & 35.6 & 48.3 & 72.3 & $>$100 \\
        \rowcolor{cyan!5}
        AEC \cite{peng2023better} & AAAI'23 & Hyper Histogram & 47.0 & 10.6 & 48.4 & 37.6 & 46.5 \\
        \rowcolor{green!5}
        GET-T \cite{peng2023get} & ICCV'23 & Group Tokens & 47.9 & 16.8 & 48.4 & 18.2 & 21.9 \\
        \rowcolor{yellow!9}
        ERGO-12 \cite{zubic2023chaos} & ICCV'23 & ERGO-12 & 50.4 & 69.9 & 40.6 & 100.0 & 59.6 \\
        \rowcolor{orange!9}
        SAST-CB \cite{peng2024scene} & CVPR'24 & Voxel Grid & 48.2 & 22.7 & 48.7 & 23.6 & 18.9 \\
        \rowcolor{brown!12}
        MAD-Det \cite{11164466} & TIP'25       & MAD        & 49.2 & 15.2 & 49.5 & 17.2 & - \\
        \rowcolor{gray!8}
        RVT-B \cite{gehrig2023recurrent} & CVPR'23 & 2D Histogram & 47.2 & 7.2 & 47.4 & 11.9 & 18.5 \\
        \rowcolor{gray!8}
        S5-ViT-B \cite{zubic2024state} & CVPR'24 & 2D Histogram & 47.4 & 8.2 & 47.2 & 9.6 & 17.5 \\
        \rowcolor{gray!8}
        SMamba \cite{yang2025smamba} & AAAI'25 & 2D Histogram & 50.4 & 22.6 & 49.3 & 26.0 & 16.7 \\
        \rowcolor{gray!8}
        SATE \cite{zhurethinking}    & NeurIPS'25 & 2D Histogram & 52.7 & 6.8 & 49.1 & 13.3 & 26.4 \\
        \rowcolor{gray!8}
        EvRT-DETR-B \cite{torbunov2025evrtdetr} & ICCV'25 & 2D Histogram & 52.7 & 7.8 & \underline{50.1} & 12.9 & 57.1 \\
        \rowcolor{blue!7}
        \hline
         Ev-DTAD-T (ours) & - & HTA & \underline{53.1} & 5.6 & 48.6 & 8.0 & 35.2 \\
        \rowcolor{blue!7}
        Ev-DTAD-B (ours) & - & HTA & \textbf{53.5} & 7.6 & \textbf{50.6} & 12.1 & 57.8 \\
        \bottomrule
    \end{tabular}
    }
\end{table*}

\subsection{Overall Ev-DTAD Framework}
\label{sec:evdtad}

Building upon HTA and FHTF, Ev-DTAD forms a complete representation-to-detection framework for EOD. 
Asynchronous events are sliced into temporal windows, converted offline into compact HTA frames, and organized into clips for video-level detection. 
An RT-DETR-style detector with a PResNet backbone and encoder extracts multi-scale features, which are refined by FHTF through recurrent temporal propagation and frequency-aware hypergraph reasoning. 
A DETR-style query decoder then predicts object categories and bounding boxes. 
Training follows a two-stage protocol: a single-frame detector is first trained on HTA frames, followed by temporal refinement training on clips. 
During temporal training, RNN states are preserved across sequential clips/videos and reset for random clips to maintain temporal consistency. 
The model is optimized with Hungarian matching and standard detection losses, including classification, $\ell_1$ box regression, and generalized IoU losses.

\section{Experiments}

\subsection{Experimental Setup}
\label{exp:setup}

\newcommand{\tabstrut}{\rule{0pt}{2.4ex}}

\noindent\textbf{Datasets.} Ev-DTAD is evaluated on three representative EOD benchmarks: Gen1~\cite{detournemire2020large}, 1Mpx/Gen4~\cite{perot2020learning}, and eTraM~\cite{verma2024etram}. Gen1 provides low-resolution automotive recordings with sparse annotations, while 1Mpx/Gen4 contains higher-resolution driving scenes with denser annotations. eTraM further evaluates static traffic monitoring under diverse lighting and weather conditions. These benchmarks enable a comprehensive evaluation across low-resolution automotive scenes, high-resolution dense traffic scenes, and static traffic monitoring scenarios, covering diverse spatial resolutions, annotation densities, and event acquisition settings. Together, they test the method across substantial changes in event density, viewpoint, and scene dynamics.

\noindent\textbf{Event Representations.} Following the standard event-to-frame protocol, we partition the asynchronous event stream into fixed 50 ms windows and convert each window into the proposed HTA representation. For Gen1, with an original resolution of $240 \times 304$, each representation is padded to $256 \times 320$ for divisibility by 32 in the feature pyramid. For 1Mpx/Gen4 and eTraM, with original resolution $720 \times 1280$, representations are downsampled to $360 \times 640$ by bilinear interpolation and padded to $384 \times 640$. This protocol follows prior work and enables fair computational comparison across datasets. Using the same window duration also provides a consistent temporal basis for evaluating different event representations. For temporal modeling, consecutive windows are grouped into clips, with length 21 for Gen1 and 10 for 1Mpx/Gen4 and eTraM.

\begin{table}[!t]
    \centering
    \caption{Comparison with state-of-the-art methods on the eTraM traffic-monitoring dataset. 
    \textcolor{red}{Runtime excludes representation generation.}
    }
    \label{tab:efficiency_comparison}
    \scriptsize
    \renewcommand{\arraystretch}{1.10}
    \setlength{\tabcolsep}{2pt}

    \begin{tabularx}{\columnwidth}{lYYYY}
        \toprule
        \textbf{Method}
        & \textbf{Representation}
        & \textbf{mAP (\%)}
        & \textbf{Runtime (ms)}
        & \textbf{Params (M)} \\
        \midrule

        \rowcolor{yellow!9}
        SAST-CB~\cite{peng2024scene}
        & Voxel Grid
        & 30.0
        & 24.4
        & 18.9 \\

        \rowcolor{gray!8}
        RVT-B~\cite{gehrig2023recurrent}
        & 2D Histogram
        & 29.5
        & 11.9
        & 18.5 \\

        \rowcolor{gray!8}
        S5-ViT-B~\cite{zubic2024state}
        & 2D Histogram
        & 29.3
        & 10.9
        & 17.5 \\

        \rowcolor{gray!8}
        SMamba~\cite{yang2025smamba}
        & 2D Histogram
        & 32.6
        & 25.2
        & \textbf{16.7} \\

        \midrule

        \rowcolor{blue!7}
        Ev-DTAD-T (ours)
        & HTA
        & \underline{35.3}
        & 8.3
        & 35.2 \\

        \rowcolor{blue!7}
        Ev-DTAD-B (ours)
        & HTA
        & \textbf{35.6}
        & 12.3
        & 57.8 \\
        \bottomrule
    \end{tabularx}
    \vspace{-2mm}
\end{table}

\noindent\textbf{Two-stage training.} Ev-DTAD is trained in two stages, with Ev-DTAD-T and Ev-DTAD-B denoting lightweight and base variants using PResNet-18 and PResNet-50 backbones, respectively. A single-frame detector is first trained for 400k iterations to learn stable spatial representations, after which the temporal refinement branch is trained for 200k iterations from the single-frame checkpoint. During temporal training, newly added temporal parameters are optimized, while the backbone, encoder, and most spatial layers remain frozen. This staged optimization preserves the learned spatial detector and isolates the learning of temporal refinement, thereby improving training stability. The total batch size is 32 for single-frame training and 8 for temporal training. All experiments use a single NVIDIA RTX 3090 GPU.

\rev{\noindent\textbf{Runtime evaluation}. Ev-DTAD inference is measured on a single NVIDIA RTX 3090 GPU with batch size 1 and FP32 precision. Representation generation in Tab.~\ref{tab:representation_comparison} is measured separately on a single thread of an Intel Xeon Gold 6336Y CPU. We report the mean per-window thread CPU time over five repeated runs after 20 warm-up windows, excluding data I/O, window partitioning, and serialization.}

\subsection{Comparison with State-of-the-Art Methods}
\label{sec:sota}

\noindent\textbf{Results on Gen1 and 1Mpx/Gen4.}
\rev{Ev-DTAD is compared with representative state-of-the-art EOD methods on Gen1 and 1Mpx/Gen4 in Tab. I. On Gen1, Ev-DTAD-B achieves the best performance of 53.5\% mAP, outperforming EvRT-DETR-B by 0.8 mAP points. Meanwhile, the lightweight Ev-DTAD-T reaches 53.1\% mAP with the fastest inference speed of 5.6 ms, maintaining accuracy close to the base variant. On 1Mpx/Gen4, Ev-DTAD-B obtains the best result of 50.6\% mAP, improving over EvRT-DETR-B by 0.5 points, while Ev-DTAD-T achieves 48.6\% mAP at the fastest runtime of 8.0 ms. The accuracy gap between the two variants thus increases from 0.4 points on Gen1 to 2.0 points on 1Mpx/Gen4, suggesting that additional backbone capacity may be more beneficial for exploiting fine object structures and multi-scale context in higher-resolution driving scenes. This complements the observation on Gen1, where the lightweight variant preserves most of the detection accuracy with lower inference latency. Compared with histogram-based detectors and methods using richer event representations, Ev-DTAD provides a competitive accuracy–efficiency trade-off across both datasets.}

\noindent\textbf{Results on eTraM.}
\rev{Tab.~\ref{tab:efficiency_comparison} reports the evaluation on eTraM, which captures static traffic monitoring scenes with varied viewpoints, object scales, and illumination conditions. Ev-DTAD-B achieves 35.6\% mAP, outperforming SMamba by 3.0 mAP points, while Ev-DTAD-T reaches 35.3\% mAP at an inference latency of 8.3 ms per frame. These results extend the evaluation beyond automotive driving scenes and demonstrate a competitive accuracy–efficiency trade-off in static traffic monitoring under dataset-specific training.}

\begin{table}[t]
    \centering
    \caption{Comparison of event representations on Gen1 using Ev-DTAD-B.
    $C$ and $t$ denote the input channels and average generation time per
    50-ms event frame, respectively.}
    \label{tab:representation_comparison}
    \scriptsize
    \setlength{\tabcolsep}{2.2pt}
    \renewcommand{\arraystretch}{1.10}

    \begin{tabularx}{0.9\columnwidth}{
        @{}
        >{\raggedright\arraybackslash}X
        >{\centering\arraybackslash}p{0.08\columnwidth}
        >{\centering\arraybackslash}p{0.14\columnwidth}
        >{\centering\arraybackslash}p{0.17\columnwidth}
        @{}
    }
        \toprule
        \textbf{Event Representation}
        & \textbf{$C$}
        & \textbf{$t$ (ms)} $\downarrow$
        & \textbf{mAP (\%)} $\uparrow$ \\
        \midrule

        Stacked 2D Histogram~\cite{torbunov2025evrtdetr}
        & 20
        & 9.83
        & 52.9 \\

        \rowcolor{gray!12}
        Time Surface~\cite{gallego2022event}
        & 2
        & 5.28
        & 53.0 \\

        ERGO-12~\cite{zubic2023chaos}
        & 12
        & 22.18
        & 52.0 \\


        \midrule
        \rowcolor{gray!12}
        HTA (Ours)
        & 3
        & 8.93
        & \textbf{53.5} \\

        \bottomrule
    \end{tabularx}
    \vspace{-2mm}
\end{table}

\subsection{Comparison with Representative Event Representations}

To assess whether the gains of HTA arise from its temporal aggregation
mechanism rather than simply from a richer input encoding, we compare it with
representative event representations under the same Ev-DTAD-B architecture and
training protocol. Only the input representation is changed across the
experiments. This controlled setting isolates the influence of representation
design from changes in detector capacity or optimization. \rev{Tab.~\ref{tab:representation_comparison}, HTA achieves the best performance of 53.5\% mAP, outperforming Stacked 2D Histogram, Time Surface, and ERGO-12 by 0.6, 0.5, and 1.5 mAP points, respectively.}

Notably, this improvement is achieved using only three input channels. In
comparison, Stacked 2D Histogram and ERGO-12 require 20 and 12 channels,
respectively, while obtaining lower accuracy and incurring higher generation
costs. Although Time Surface is faster and uses two channels, its detection
performance remains 0.5 points below HTA. These results indicate that the
advantage of HTA does not stem from increasing the representation dimensionality.
Instead, its hierarchical aggregation preserves detection-relevant temporal
information in a compact form, providing a favorable balance between
representation quality and generation efficiency.

\rev{The generation times in Tab.~\ref{tab:efficiency_comparison} are reported separately from detector inference. On Gen1, HTA requires 8.93 ms per 50-ms event window, compared with 9.83 ms for Stacked 2D Histogram and 5.28 ms for Time Surface. HTA inputs are precomputed in our evaluation, so this preprocessing cost is excluded from the Ev-DTAD runtimes in Tabs~\ref{tab:main_results} and ~\ref{tab:efficiency_comparison}. }

\begin{table}[!t]
    \centering
    \caption{Ablation study on the individual and joint contributions of HTA and FHTF.}
    \label{tab:ablation_representation}
    \scriptsize
    \setlength{\tabcolsep}{3pt}
    \renewcommand{\arraystretch}{1.12}

    \begin{tabularx}{\columnwidth}{YYYYY}
        \toprule
        \textbf{HTA}
        & \textbf{FHTF}
        & \textbf{mAP (\%)}
        & \textbf{mAP$_{50}$ (\%)}
        & \textbf{mAP$_{75}$ (\%)} \\
        \midrule

        \textcolor{evred}{\xmark}
        & \textcolor{evred}{\xmark}
        & 47.6
        & 75.6
        & 49.5 \\

        \rowcolor{gray!12}
        \textcolor{evgreen}{\cmark}
        & \textcolor{evred}{\xmark}
        & 51.3
        & 79.7
        & 53.8 \\

        \textcolor{evred}{\xmark}
        & \textcolor{evgreen}{\cmark}
        & 52.9
        & 82.2
        & 55.9 \\

        \rowcolor{gray!12}
        \textcolor{evgreen}{\cmark}
        & \textcolor{evgreen}{\cmark}
        & 53.5
        & 82.2
        & 56.7 \\
        \bottomrule
    \end{tabularx}

    \vspace{-2mm}
\end{table}

\subsection{Ablation Study}
\label{sec:ablation}

Unless otherwise specified, all ablation studies and component analyses in Sec. ~\ref{sec:ablation} and Sec. ~\ref{sec:further_analysis} are conducted on the Gen1 benchmark under the same evaluation protocol as the main comparison.

\noindent\textbf{Effectiveness of HTA and FHTF.}
HTA and FHTF are first ablated in Tab.~\ref{tab:ablation_representation}. Compared with the 2D-HG baseline without FHTF, HTA brings a +3.7 mAP improvement, showing the benefit of compact representation-level temporal encoding. FHTF brings a larger +5.3 mAP improvement over the same baseline, indicating the effectiveness of temporal-relational feature modeling. \rev{Combining HTA and FHTF achieves the best result of 53.5\% mAP and improves mAP75 by 7.2 points over the baseline. In particular, introducing FHTF on top of HTA further increases mAP from 51.3\% to 53.5\%. This additional gain is consistent with their different roles in temporal modeling: HTA preserves event-response history through temporal weighting and reliability-aware accumulation before feature extraction, while FHTF learns to propagate detection-relevant feature states across consecutive representations and refines their high-order associations. These results support the complementary benefit of combining temporally enriched inputs with task-adaptive temporal-relational refinement in the feature space.}

\begin{table}[!t]
    \centering
    \caption{Ablation study on intra- and inter-window aggregation in HTA.}
    \label{tab:ablation_hta_rgb}
    \scriptsize
    \setlength{\tabcolsep}{3pt}
    \renewcommand{\arraystretch}{1.10}

    \begin{tabularx}{0.8\columnwidth}{YYYY}
        \toprule
        \textbf{Intra-Agg.}
        & \textbf{Inter-Agg.}
        & \textbf{mAP (\%)}
        & \textbf{mAP$_{50}$ (\%)} \\
        \midrule

        \textcolor{evred}{\xmark}
        & \textcolor{evred}{\xmark}
        & 51.3
        & 79.5 \\

        \rowcolor{gray!12}
        \textcolor{evgreen}{\cmark}
        & \textcolor{evred}{\xmark}
        & 51.7
        & 80.2 \\

        \textcolor{evred}{\xmark}
        & \textcolor{evgreen}{\cmark}
        & 53.2
        & 82.1 \\

        \rowcolor{gray!12}
        \textcolor{evgreen}{\cmark}
        & \textcolor{evgreen}{\cmark}
        & 53.5
        & 82.2 \\
        \bottomrule
    \end{tabularx}

    \vspace{-2mm}
\end{table}

\begin{table}[!t]
    \centering
    \caption{Ablation study of the FHTF module, including FERSH and ConvLSTM.}
    \label{tab:ablation_FHTF}
    \scriptsize
    \setlength{\tabcolsep}{2.5pt}
    \renewcommand{\arraystretch}{1.10}

    \begin{tabularx}{0.9\columnwidth}{YYYYY}
        \toprule
        \textbf{FERSH}
        & \textbf{ConvLSTM}
        & \textbf{mAP (\%)}
        & \textbf{Runtime (ms)}
        & \textbf{Params (M)} \\
        \midrule

        \textcolor{evred}{\xmark}
        & \textcolor{evred}{\xmark}
        & 51.3
        & 6.2
        & 42.8 \\

        \rowcolor{gray!12}
        \textcolor{evgreen}{\cmark}
        & \textcolor{evred}{\xmark}
        & 51.8
        & 7.2
        & 43.5 \\

        \textcolor{evred}{\xmark}
        & \textcolor{evgreen}{\cmark}
        & 52.5
        & 7.3
        & 57.1 \\

        \rowcolor{gray!12}
        \textcolor{evgreen}{\cmark}
        & \textcolor{evgreen}{\cmark}
        & 53.5
        & 7.6
        & 57.8 \\
        \bottomrule
    \end{tabularx}

    \vspace{-2mm}
\end{table}

\noindent\textbf{Ablation on intra- and inter-window aggregation.}
Tab.~\ref{tab:ablation_hta_rgb} analyzes the two aggregation levels in HTA. Compared with the variant without hierarchical aggregation, intra-window aggregation brings a +0.4 mAP improvement, while inter-window aggregation brings a +1.9 mAP improvement. Combining both achieves the best performance of 53.5\% mAP and improves the no-aggregation variant by +2.2 mAP. These results show that intra-window aggregation captures local temporal evidence, while inter-window aggregation preserves temporal continuity across adjacent windows.

\noindent\textbf{Ablation on FHTF components.}
Tab.~\ref{tab:ablation_FHTF} studies FERSH and ConvLSTM within FHTF. Starting from the HTA baseline without FHTF, FERSH brings a +0.5 mAP improvement, suggesting that frequency-aware hypergraph reasoning provides useful relational refinement. ConvLSTM brings a larger +1.2 mAP improvement, showing the importance of stateful temporal modeling. Combining FERSH and ConvLSTM achieves the best result of 53.5\% mAP, indicating that high-order relational reasoning complements recurrent temporal aggregation.

\begin{figure*}[t]
  \centering
  \includegraphics[width=\textwidth]{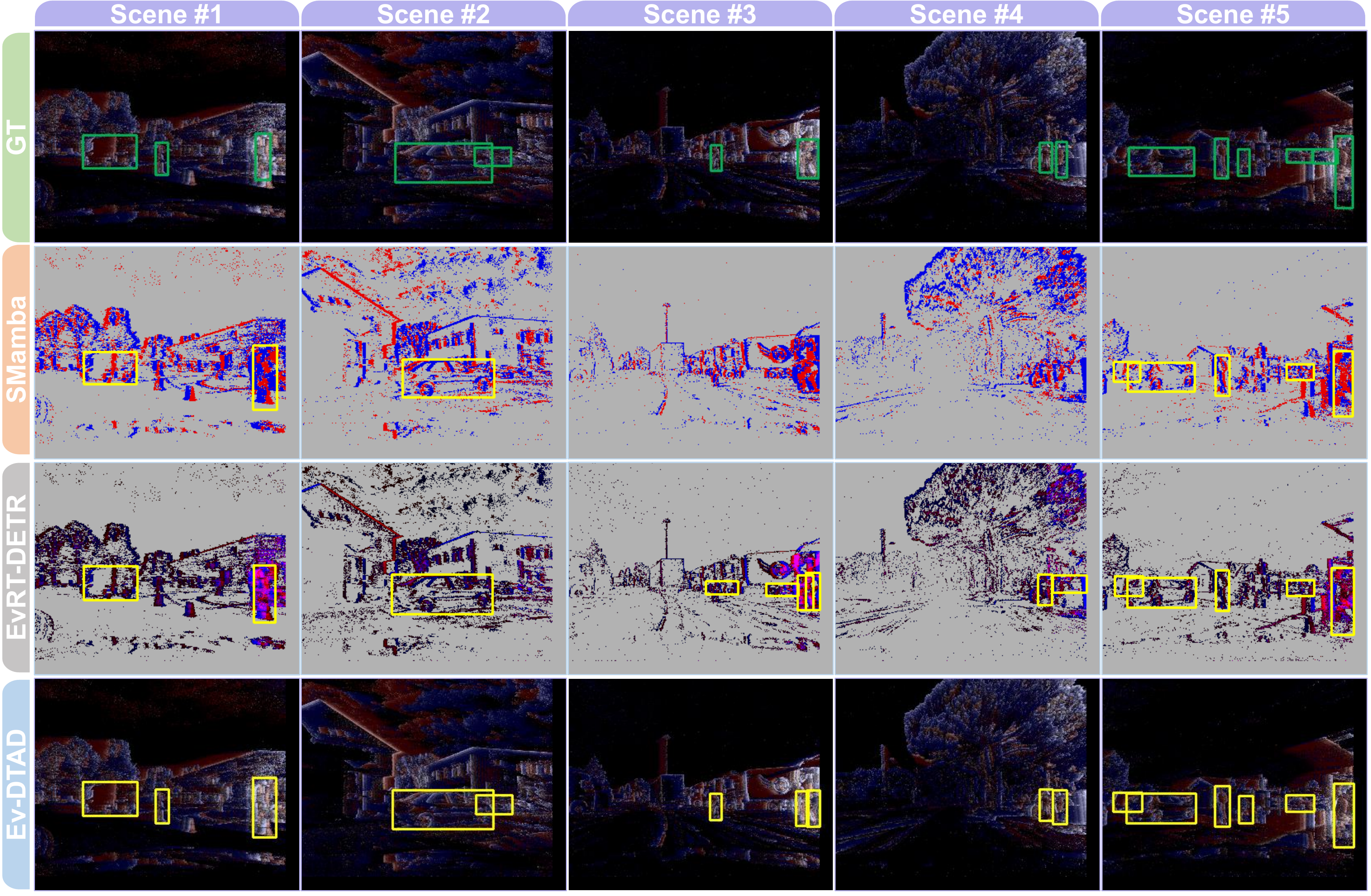}
    \caption{\textbf{Qualitative comparisons on Gen1.} Ev-DTAD achieves more accurate and robust detections than the baseline, reducing missed detections and improving localization in challenging scenes.}
  \label{fig:result_vis}
\end{figure*}

\begin{table}[!t]
    \centering
    \caption{Evaluation of the generalizability and effectiveness of HTA across different object detection architectures.}
    \label{tab:generalizability_baseline}
    \scriptsize
    \setlength{\tabcolsep}{2.5pt}
    \renewcommand{\arraystretch}{1.10}

    \begin{tabularx}{0.9\columnwidth}{
        >{\raggedright\arraybackslash
          \hsize=1.45\hsize\linewidth=\hsize}X
        >{\centering\arraybackslash
          \hsize=1.55\hsize\linewidth=\hsize}X
        >{\centering\arraybackslash
          \hsize=0.65\hsize\linewidth=\hsize}X
        >{\centering\arraybackslash
          \hsize=0.65\hsize\linewidth=\hsize}X
        >{\centering\arraybackslash
          \hsize=0.70\hsize\linewidth=\hsize}X
    }
        \toprule
        \textbf{Method}
        & \textbf{Backbone}
        & \textbf{2D-HG}
        & \textbf{HTA}
        & \textbf{$\Delta$mAP} \\
        \midrule

        RT-DETR-B~\cite{zhao2024detrs}
        & DETR
        & 47.6
        & 51.3
        & +3.7 \\

        \rowcolor{gray!12}
        D-FINE~\cite{peng2024d}
        & DETR
        & 47.6
        & 52.2
        & +4.6 \\

        RVT-B~\cite{gehrig2023recurrent}
        & Transformer+RNN
        & 47.2
        & 47.7
        & +0.5 \\

        \rowcolor{gray!12}
        Ev-DTAD-B (ours)
        & DETR+RNN
        & 52.9
        & 53.5
        & +0.6 \\
        \bottomrule
    \end{tabularx}

    \vspace{-2mm}
\end{table}

\subsection{Further Analysis}
\label{sec:further_analysis}

\noindent\textbf{Generalizability of HTA.}
To examine whether HTA is architecture-specific, it is evaluated across different object detection architectures in Tab.~\ref{tab:generalizability_baseline}. The evaluated models include mainstream image-based detectors, i.e., RT-DETR-B~\cite{zhao2024detrs} and D-FINE~\cite{peng2024d}, as well as the representative event-based detector RVT-B~\cite{gehrig2023recurrent}. Compared with the standard 2D-HG representation, HTA consistently improves detection performance across all architectures. In particular, it brings clear gains on mainstream DETR-based detectors, improving RT-DETR-B from 47.6\% to 51.3\% mAP and D-FINE from 47.6\% to 52.2\% mAP. This indicates that HTA can serve as a general event representation rather than a detector-specific preprocessing design.

\begin{figure}[!t]
    \centering
    \includegraphics[
        width=0.88\columnwidth
    ]{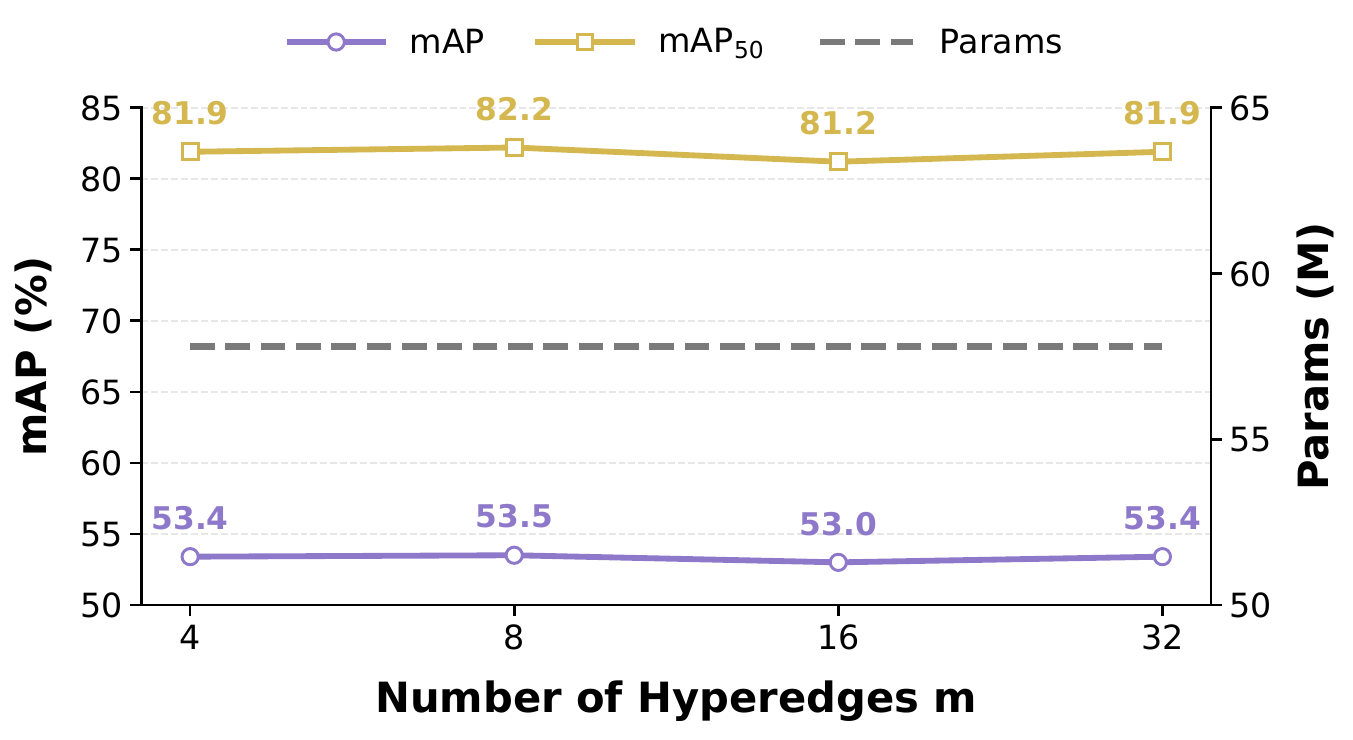}
    \caption{Hyperedge sensitivity analysis of FHTF module.}
    \label{fig:ablation_hyperedgenum}
\end{figure}

\rev{We further observe relatively modest improvements when recurrent modeling is already present, with gains of 0.5 and 0.6 mAP points on RVT-B and Ev-DTAD-B, respectively. This pattern may reflect partial functional overlap between the event history encoded by HTA and the temporal context aggregated by recurrent modules. Nevertheless, the consistent gains across the evaluated architectures indicate that HTA remains beneficial even when recurrent modeling is employed. Since Tab.~\ref{tab:generalizability_baseline} evaluates changes in input representation while retaining each detector architecture, these results support the applicability of HTA across architectures but do not establish that recurrent modules can be replaced.}

\noindent\textbf{Hyperedge sensitivity.}
\rev{Fig~\ref{fig:ablation_hyperedgenum} analyzes the sensitivity of FHTF to the number of hyperedges $m$. Across the evaluated settings ($m=4,8,16,32$), detection performance remains stable, with mAP ranging from 53.0\% to 53.5\% and mAP50 from 81.2\% to 82.2\%, while all configurations contain approximately 57.8M parameters. These consistently strong results demonstrate the robustness of our method to variations in the hyperedge number. We adopt $m=8$ as the default setting because it achieves the highest observed mAP and mAP50 with a compact hyperedge set, whereas increasing the number to 16 or 32 provides no further accuracy gains.}

\begin{figure*}[t]
  \centering
  \includegraphics[width=\textwidth]{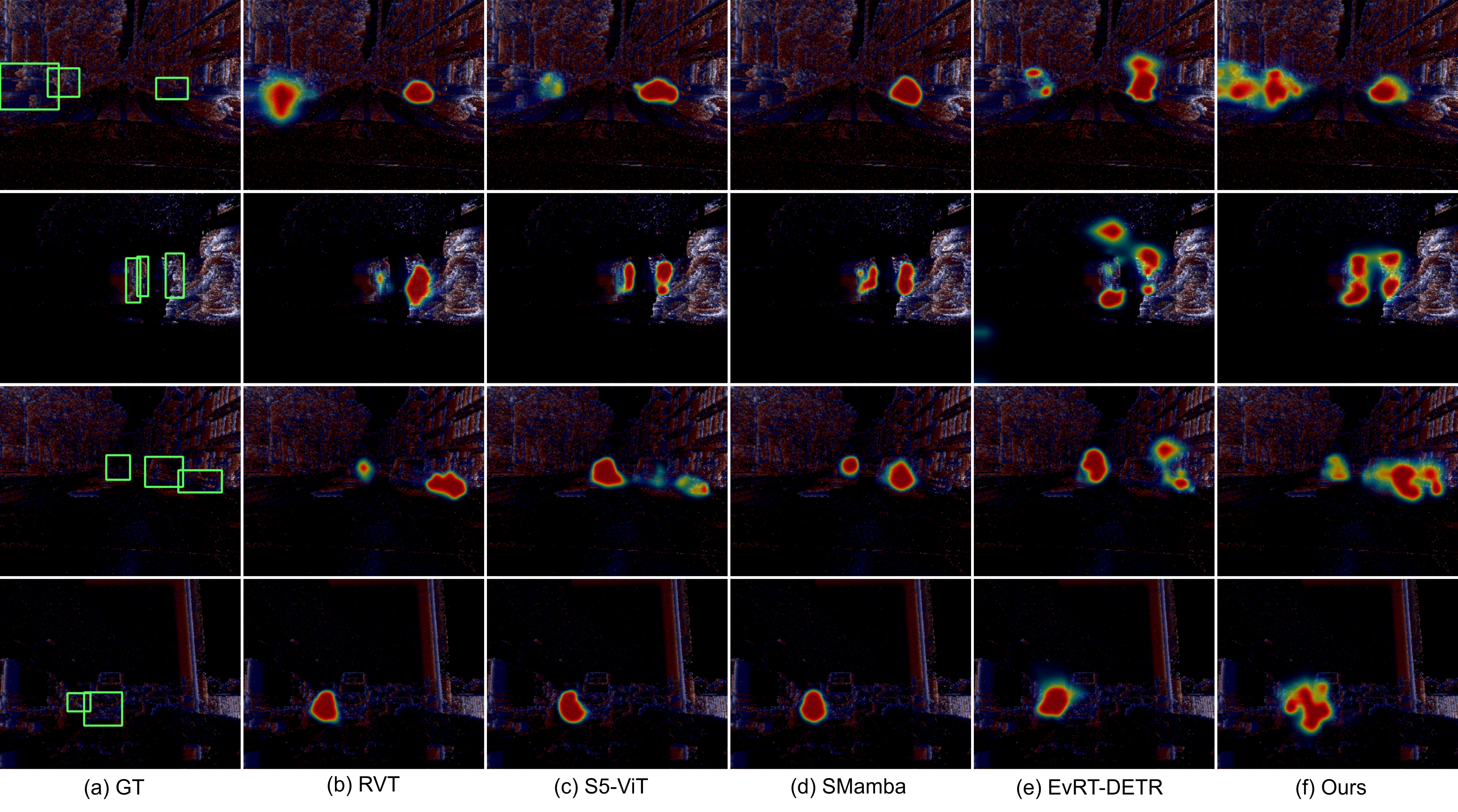}
    \caption{\textbf{Qualitative heatmap comparison.} (a) Ground truth; (b)--(f) RVT, S5-ViT, SMamba, EvRT-DETR, and Ev-DTAD (ours). Our method more consistently highlights object regions while suppressing background responses.}
  \label{fig:heatmap}
\end{figure*}

\subsection{Qualitative Analysis}
\label{sec:qualitative_analysis}

\noindent\textbf{Detection results.}
Fig.~\ref{fig:result_vis} compares Ev-DTAD with SMamba and EvRT-DETR across five representative Gen1 scenes. Existing methods exhibit missed detections or inaccurate localization when objects are small, spatially adjacent, or represented by sparse event responses. In contrast, Ev-DTAD detects more ground-truth instances and produces bounding boxes that align more closely with the object regions, particularly in Scenes~\#1, \#3, and \#4. It also maintains stable detection in the relatively crowded Scene~\#5. These qualitative results demonstrate that the proposed temporal aggregation and feature refinement improve both object recognition and localization under challenging event patterns.

\noindent\textbf{Activation heatmaps.}
\rev{Fig.~\ref{fig:heatmap} compares detection-guided LayerCAM responses on temporally aligned Gen1 frames. Features at strides 8, 16, and 32 are extracted from the PAFPN outputs of RVT-B, S5-ViT-B, and SMamba-Base, and from the temporal encoder outputs preceding the decoders of EvRT-DETR-P18 and Ev-DTAD. LayerCAM aggregates channels using positive gradients of the summed matched prediction scores. The resulting maps are resized and cropped, normalized by their spatial maxima, and averaged across scales. Each fused map is independently clipped and rescaled using its 5th and 99th percentiles, with identical contrast adjustment, smoothing, and color mapping across models. In these examples, Ev-DTAD exhibits more concentrated responses around annotated objects. This comparison concerns relative spatial response patterns rather than absolute activation magnitudes across models.}



\section{Conclusion}

In this paper, we present Ev-DTAD, a unified EOD framework that addresses temporal modeling at two complementary levels: representation-level temporal event encoding and model-level temporal-hypergraph reasoning. Specifically, HTA integrates intra-window temporal ordering with reliable inter-window temporal propagation into a compact three-channel pseudo-RGB representation. FHTF further refines multi-scale event features by jointly modeling temporal evolution and frequency-aware high-order dependencies, thereby aggregating sparse and fragmented event responses into more coherent object features. Experiments on Gen1 (+0.8 mAP), 1Mpx/Gen4 (+0.5 mAP), and eTraM (+3.0 mAP) demonstrate a competitive accuracy--efficiency trade-off across automotive driving and static traffic monitoring scenarios. Controlled comparisons with representative event representations further show that HTA achieves the highest mAP using only three input channels, confirming that its advantage arises from effective temporal organization rather than increased representation dimensionality. Moreover, the consistent improvements across different image-based and event-specific detection architectures validate the generalizability of HTA beyond Ev-DTAD. Overall, these findings demonstrate the complementarity between compact temporal representation and feature-level temporal-hypergraph reasoning, and suggest that stronger representation-level temporal encoding can reduce the burden placed solely on heavy recurrent modeling, providing a promising direction for accurate and efficient EOD systems.

\bibliographystyle{IEEEtran}
\bibliography{ref}

\end{document}